\documentclass[10pt]{article}
\usepackage{color, palatino}

\usepackage[psamsfonts]{amssymb}
\usepackage{amsmath, amsthm, graphicx, enumerate, subfigure, color,bbm}
\usepackage{amsfonts}
\usepackage{color}
\usepackage[top=3cm, bottom=3cm, left=3.5cm, right=3.5cm]{geometry}
\usepackage{algorithmic}
\usepackage{hyperref}
\usepackage[all]{hypcap}   
\usepackage{enumerate}
\usepackage{natbib}
\usepackage{lipsum}
\hypersetup{
    bookmarks=true,         
    unicode=false,          
    pdftoolbar=true,        
    pdfmenubar=true,        
    pdffitwindow=false,     
    pdfstartview={FitH},    
    pdftitle={My title},    
    pdfauthor={Author},     
    pdfsubject={Subject},   
    pdfcreator={Creator},   
    pdfproducer={Producer}, 
    pdfkeywords={keyword1} {key2} {key3}, 
    pdfborder={0 0 2 [3 3]},
    pdfnewwindow=true,      
    colorlinks=true,       
    linkcolor=red,          
    citecolor=green,        
    filecolor=magenta,      
    urlcolor=cyan           
}

\usepackage{booktabs}
\usepackage{array} 
\usepackage{paralist}  
\usepackage{verbatim} 
\usepackage{booktabs}
\usepackage{multirow}
\usepackage{subfigure} 
\usepackage{amsmath}
\usepackage{listings}
\usepackage{rotating}
\usepackage{hyperref}
\hypersetup{
	colorlinks,%
	citecolor=black,%
	filecolor=black,%
	linkcolor=black,%
	urlcolor=black
}
\usepackage[ruled,noline]{algorithm2e}
\newcommand{\pp}{{$^{++}$}}

\begin{document}

\title{GURLS: a Least Squares Library for  Supervised Learning}

\author{Andrea Tacchetti$^\diamond$$^\S$ , Pavan K.~Mallapragada$^\diamond$, Matteo~Santoro$^\S$, Lorenzo~Rosasco$^\S$ \\
\small \it    $^\diamond$ Center for Biological and Computational Learning, Massachusetts Institute of Technology, Cambridge (MA) USA \\
\small \it        $^\S$Laboratory  for Computational and Statistical Learning,  Istituto Italiano di Tecnologia,  Genova, Italy\\
\small \tt  \{atacchet, pavan\_m, lrosasco\}@mit.edu, matteo.santoro@iit.it}

\maketitle

\begin{abstract}
We present GURLS, a least squares,  modular,  easy-to-extend software library for efficient supervised learning.  GURLS is targeted  to machine learning practitioners, as well as non-specialists. 
It  offers a number state-of-the-art training strategies for medium and large-scale learning, 
and routines for  efficient  model selection.  The library  is particularly well suited for  multi-output problems (multi-category/multi-label).
GURLS  is currently available in two independent implementations: Matlab and C++.  It takes advantage of the  favorable properties of regularized least squares algorithm  to  exploit advanced tools in linear algebra. Routines to  handle computations with very large matrices     by means of memory-mapped storage and distributed task execution are available. 
 The package is distributed under the BSD licence and is available for download at~\texttt{https://github.com/CBCL/GURLS}.  
\end{abstract}

\section{Introduction and Design}

Supervised learning has become a fundamental tool for the design of intelligent systems and the analysis of complex high dimensional data.  Key to the success of supervised learning has been the availability of efficient, easy-to-use software packages. Novel data collection technologies make it easy to gather  high dimensional, multi-output data sets of  ever  increasing size (Big Data). This fact  calls for new software solutions for  the automatic  
training, tuning and testing of  effective and efficient supervised learning methods.

These observations motivate the design of GURLS (which stands for Grand Unified Regularized Least Squares). 
Specifically, the package was developed to pursue the following goals: 
{\it Speed}: Fast training/testing  procedures (online, batch, randomized, distributed) for learning problems with potentially large/huge number of points,  features and especially outputs (e.g. classes).  
{\it Memory}: Flexible data management to work with large datasets   by means of memory-mapped storage.
{\it Performance}:  State of  the art results in high-dimensional multi-output  problems (e.g. object recognition tasks  with  tens or hundreds of classes, where the input have dense features). 
{\it Usability and modularity}: Easy to use and to expand library.

GURLS is based on Regularized Least Squares (RLS) and takes advantage of all the favorable properties of these methods~\citep{rifkin2003regularized}.  First, and foremost, since  the algorithm reduces to solving a linear system, GURLS is set up to 
exploit the powerful tools,  and recent advances, of linear algebra (including randomized solver, 
first order methods, etc.). Second, it makes use of  RLS properties which are particularly suited for 
high dimensional learning. For example:  
$(1)$ RLS has natural primal and dual formulation (hence having complexity which is the smallest between number of examples and features); 
$(2)$ efficient parameter selection (closed form expression of the leave one out error and efficient computations of regularization path); 
$(3)$ natural and efficient extension to multiple outputs.  
Specific attention has been devoted to handle large high dimensional ({\em Big}) data. Indeed, we rely on data structures that can be serialized using memory-mapped files, and on a distributed task manager to perform a number of key steps (such as matrix multiplication) without loading the whole dataset in memory.

Specific attention has been taken to provide a lean API and an exhaustive  documentation. 
GURLS has minimal  external dependencies, and has been deployed and tested successfully 
on Linux, MacOS and Windows. The library is distributed under the simplified BSD license, and can be downloaded from~\texttt{https://github.com/CBCL/GURLS}.

\section{Description of the library}\label{sec:software}

The library comprises four main modules. GURLS and bGURLS -- both implemented in Matlab -- are aimed at solving learning problems with small/medium and 
large-scale datasets respectively. GURLS\pp and bGURLS\pp are their C++ counterparts. 
They share the same design of the former two modules, but have significant 
improvements made possible by the use of C++, which makes them faster and more flexible.

The specification of the desired machine learning experiment in the library is very straightforward. Basically, it is a formal description of a {\em pipeline}, i.e. an ordered 
sequence of steps. Each step identifies an actual learning task, which can belong to a predefined category. The core of the library is a method (a class in the C\pp 
implementation) called {\it GURLScore}, which is responsible for processing the sequence of tasks in the proper order and for linking the output of the former task to the 
input of the subsequent one. 
A key role is played by the additional ``options'' structure, which we usually refer to as {\sc opt}. It is used to store all configuration parameters required to customize the 
behavior of individual tasks in the pipeline. 
Tasks receive configuration parameters from {\sc opt} in read-only mode and -- upon termination -- the results are appended to the 
structure by {\it GURLScore} in order to make them available to the subsequent tasks. This allows the user to easily skip the execution of some tasks in a pipeline, by 
simply inserting the desired results directly into the options structure. 
Currently, we identify six different task categories: dataset splitting, kernel computation, model selection, training, evaluation and testing, performance assessment and analysis. 
Tasks belonging to the same category may be interchanged with each other.  

\subsection{Learning from large datasets}

Two modules in GURLS have been specifically designed to deal with {\em big learning} scenarios. The approach we adopted is mainly based on a memory-
mapped abstraction of matrix and vector data structures, and on a distributed computation of a number of standard problems in linear algebra. Without the ambition to 
develop a good solution for all the possible variants of big learning, we decided to focus  specifically on those situation where one seeks a linear model on a large 
set of (possibly non linear) features. A more accurate specification of what ``large'' means in GURLS is directly related to the number of features ($d$) and the number of training examples ($n$): we require it must 
be possible to store a $\min(d,n)\times \min(d,n)$ matrix in memory. 
In practice, this roughly means we can train models with up-to $25k$ features on machines with $8Gb$ of RAM, and 
up-to $50k$ features on machines with $36Gb$ of RAM. It is important to remark we {\em do not} require the data matrix itself to be stored in memory. 
Indeed, in GURLS it is possible to manage an arbitrarily large set of training examples. 

We distinguish two different scenarios. Data sets that can fully reside in RAM without any memory mapping techniques -- such as swapping -- are considered to be small/
medium. Larger data sets are considered to be ``big'' and learning must be performed using either bGURLS or bGURLS\pp. These two modules include all the design 
patterns described above, and have been complemented with additional big data and distributed computation capabilities. Big data support is obtained using a data 
structure called {\it bigarray}, which allows to handle data matrices as large as a machine's available space on hard drive instead of its RAM: we store the entire dataset on 
disk and load only small chunks in memory when required. Due to programming language constraints, there are some differences between the Matlab and C\pp implementations.

bGURLS relies on a simple interface -- developed ad-hoc and called {\it GURLS Distributed Manager} (GDM) -- to distribute matrix-matrix multiplications, thus allowing users to perform the important task of kernel matrix computation on a distributed network of computing nodes. After this step, the subsequent tasks behave as in GURLS.

bGURLS\pp (currently in active development) offers more interesting features because it is based on the MPI libraries. Therefore, it allows for a full distribution within every single task of the pipeline. All the processes read the input data from a shared filesystem over the network and then start executing the same pipeline. During execution, each process' task communicates with the corresponding ones running over the other processes. Every process maintains his local copy of the options. Once the same task is completed by all processes, the local copies of the options are synchronized. This advanced architecture allows for the creation of hybrid pipelines comprising serial one-process-based tasks from GURLS\pp.

\section{Experiments}\label{SecExperiments}

Due to space requirements we decided to focus the experimental analysis in the paper to the assessment of GURLS' performance both in terms of accuracy and time. In our experiments we considered 5 popular data sets, briefly described in Table~\ref{tab:datasets}. Experiments were run on a Intel Xeon 5140 @ 2.33GHz processor with  8GB of RAM, and operating system Ubuntu 8.10 Server (64 bit).

\begin{table}[t]
\begin{center}
\scriptsize
\begin{tabular}{|c|c|c|c|}
\hline
 & \# of & \# of & \# of\\
data set &  samples & classes & variables\\
\hline
optdigit & 3800 & 10 & 64\\
\hline
landast & 4400 & 6 & 36\\
\hline
pendigit & 7400 & 10 & 16\\
\hline
letter & 10000 & 26 & 16\\
\hline
isolet & 6200 & 26 & 600\\
\hline
\end{tabular}
\caption{Data sets description.}\label{tab:datasets}
\end{center}
\end{table}

\begin{table}[!ht]
\begin{center}\scriptsize
\begin{tabular}{|c|cc|cc|cc|} 
\hline
& \bf optdigit& & \bf landsat & & \bf pendigit & \\
& accuracy (\%) & time (s)& accuracy (\%) &time (s)&accuracy (\%) &time (s)\\
\hline
\hline
GURLS (linear primal) &92.3 & 0.49 & 63.68 & 0.22 & 82.24 & 0.23 \\
\hline
GURLS (linear dual)  &92.3 & 726 & 66.3 & 1148 & 82.46 & 5590 \\
\hline
LS-SVM linear & 92.3 & 7190 & 64.6 & 6526 & 82.3 & 46240 \\
\hline
GURLS (500 random features) & 96.8 &  25.6 & 63.5& 28.0 & 96.7 & 31.6 \\
\hline
GURLS (1000 random features) & 97.5 & 207 &  63.5 & 187 & 95.8 & 199\\
\hline
GURLS (gaussian kernel) & 98.3 &  13500 & 90.4& 20796 & 98.4 & 100600\\
\hline
LS-SVM (gaussian kernel) & 98.3 & 26100 & 90.51& 18430 & 98.36 & 120170 \\
\hline
\end{tabular}
\caption{Comparison between GURLS and LS-SVM.}\label{tab:exp-lssvm}
\end{center}
 \end{table}

\begin{figure}[!ht]\centering
\includegraphics[width=\linewidth]{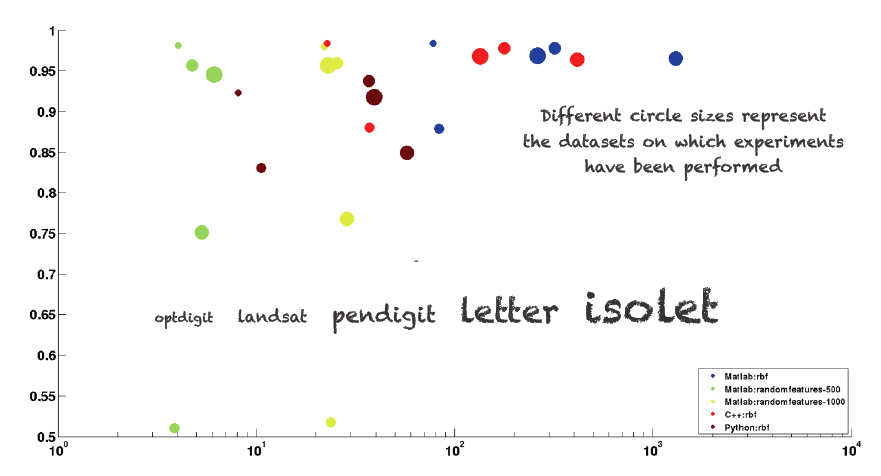}
\caption{Prediction accuracy (from $0.5$ to $1$) vs. computing time (in seconds). The color of the circles represents the training method and the library used. In blue we report the Matlab implementation of RLS with RBF kernel, while in red we report its C++ counterpart. In dark red are the results with the LIBSVM library with RBF kernel. Finally, in yellow and green we report the results obtained using a linear kernel on $500$ and $1000$ random features respectively.}\label{fig:exp}
\end{figure}

We set up different pipelines with different optimization routines available in GURLS, and compared the performance to SVM, for which we used the python modular interface to LIBSVM \citep{chang2001libsvm}. Automatic selection of the optimal regularization parameter is implemented identically in all experiments: $(i)$ split the data; $(ii)$ define a set of regularization parameter on a regular grid; $(iii)$ perform hold-out validation. The variance of the Gaussian kernel has been fixed by looking at the statistics of the pairwise distances among training examples. The prediction accuracy of GURLS and GURLS\pp is identical  -- as expected -- but the implementation in C++ is significantly faster. The prediction accuracy of standard RLS-based methods is in many cases higher than SVM. However the computing performance seem to favor SVM. By exploiting the wide range of optimization procedures handled in our library, we further run the experiments with the random features approximation~\citep{rahimi2008} implemented in GURLS. As show in Figure~\ref{fig:exp}, the performance of such method is comparable to that of SVM at much lower computational cost in the majority of the tested data sets.

We further compared GURLS with another available least squares based toolbox, namely the LS-SVM toolbox~\citep{LSSVM}, which includes  optimized routines for parameter selection such as {\em coupled simulated annealing} and line/grid search. The goal of this experiment is to benchmark the performance of parameter selection with random data splitting included in GURLS, which is basically an exhaustive grid search over the parameters.
For a fair comparison, we considered only the Matlab implementation of GURLS. Results are reported in Table~\ref{tab:exp-lssvm}. 
As  expected, using the linear kernel with the primal formulation -- not available in LS-SVM -- is the fastest approach since it leverages the lower dimensionality of the input space.  When the gaussian kernel is used, both GURLS and LS-SVM have comparable computing time and classification performance. Note, however, that in GURLS the number of  parameter in the grid search is fixed to $400$, while in LS-SVM it may vary significantly and is limited to $70$. We emphasize the fairly acceptable classification performance of the fast random features implementation in GURLS, which makes it a valuable choice in many applications. 
Finally, we note that  all GURLS pipelines,  in their Matlab implementation,  are generally faster than LS-SVM, and further improvements are achieved if GURLS\pp is considered.

\section*{Acknowledgements}
\small We thank Tomaso Poggio, Zak Stone, Nicolas Pinto, Hristo S. Paskov and CBCL for useful comments and insights.
\normalsize

\bibliographystyle{apalike}
\bibliography{refsgurls}

\end{document}